\documentclass[conference]{IEEEtran}
\IEEEoverridecommandlockouts
\usepackage{cite}
\usepackage{amsmath,amssymb,amsfonts}
\usepackage{algorithmic}
\usepackage{graphicx}
\usepackage{textcomp}
\usepackage{xcolor}
\def\BibTeX{{\rm B\kern-.05em{\sc i\kern-.025em b}\kern-.08em
    T\kern-.1667em\lower.7ex\hbox{E}\kern-.125emX}}
    
\newcommand{\vect}[1]{\mathbf{#1}}

\newcommand{\ra}[1]{\renewcommand{\arraystretch}{#1}}

\usepackage{pifont}
\usepackage{graphics}
\usepackage{hyperref}
\usepackage{booktabs}
\usepackage{caption, subcaption}

\begin{document}

\title{Anomaly localization for copy detection patterns through print estimations\\
\thanks{S. Voloshynovskiy is a corresponding author.}
\thanks{This research was partially funded by the Swiss National Science Foundation SNF No. 200021\_182063.}
}

\author{\IEEEauthorblockN{Brian Pulfer, Yury Belousov, Joakim Tutt, Roman Chaban, Olga Taran, Taras Holotyak and Slava Voloshynovskiy} 
\IEEEauthorblockA{\textit{Department of Computer Science, University of Geneva, Switzerland} \\
\{brian.pulfer, yury.belousov, joakim.tutt, roman.chaban, olga.taran, taras.holotyak, svolos\}@unige.ch}}

\maketitle

\begin{abstract}
Copy detection patterns (CDP) are recent technologies for protecting products from counterfeiting. However, in contrast to traditional copy fakes, deep learning-based fakes have shown to be hardly distinguishable from originals by traditional authentication systems. Systems based on classical supervised learning and digital templates assume knowledge of fake CDP at training time and cannot generalize to unseen types of fakes. Authentication based on printed copies of originals is an alternative that yields better results even for unseen fakes and simple authentication metrics but comes at the impractical cost of acquisition and storage of printed copies. In this work, to overcome these shortcomings, we design a machine learning (ML) based authentication system that only requires digital templates and printed original CDP for training, whereas authentication is based solely on digital templates, which are used to estimate original printed codes. The obtained results show that the proposed system can efficiently authenticate original and detect fake CDP by accurately locating the anomalies in the fake CDP. The empirical evaluation of the authentication system under investigation is performed on the original and ML-based fakes CDP printed on two industrial printers \footnote{Source code is publicly available at \url{https://gitlab.unige.ch/Brian.Pulfer/cdp_unsupervised_ad}}.
\end{abstract}

\begin{IEEEkeywords}
copy detection patterns, anomaly localization, anomaly detection, unsupervised deep learning.
\end{IEEEkeywords}

\section{Introduction}\label{sec:intro}
Counterfeiting hits many segments of industry. The market is nowadays affected, among others, by counterfeits of pharmaceutical medicines, luxury products, and food, as well as banknotes and even identification documents. Copy detection patterns (CDP)\cite{picard, voloshynovskiy2016physical} are a popular technique for protecting products against counterfeiting, which is a major threat to modern economy. They consist of two-dimensional digital binary codes which are printed using some industrial printers to obtain the respective printed codes, which are distributed in public domain with the associated products. When authenticating a product, the associated printed CDP is compared either with the digital template or with a printed template held by the product owner, hereinafter referred to as defender. The verification stage can be carried out with smartphones so that customers can verify the authenticity of a product directly.

Traditionally, the counterfeiting pipeline includes an enrollment of the publicly available printed original codes by the attacker by using high-resolution  scanners or special cameras followed by some hand-crafted (HC) post-processing and reprinting of the estimated codes on counterfeited products. Depending on the symbols' size and printing resolution used by the defender, CDP cannot be cloned perfectly because of the phenomenon known as dot gain. Detection of these traditional fakes is relatively trivial.

At the same time, recent works \cite{Taran2020icassp, romanwifs21, khermaza2021can} have shown that machine learning (ML) based attacks can produce high-quality copies of CDP (fakes). These attacks use the original printed templates to obtain an estimate of the respective digital templates, which are then used to print fake copies of the printed templates even using the same printer used for originals. Follow-up work \cite{icip22} has shown that classical supervised-learning authentication is susceptible to the phenomenon of distribution shift, meaning that supervised systems perform poorly in face of unseen fakes, which is often the case in real-world scenarios. Therefore, there is a high need for CDP authentication systems to be capable of reliably distinguishing the original CDP from ML-based fakes of different types without knowing these fakes in advance at the training stage.

In \autoref{sec:t_vs_x}, we show that the authentication based on printed templates performs better than authentication based on digital templates. However, holding such printed templates is impractical from the standpoint of the defender for the following reasons:

\begin{itemize}
    \item the acquisition of printed templates is time-consuming and expensive;
    \item a mismatch between enrollments taken by the defender using high-resolution cameras and those taken by the verifier's mobile phones might be significant;
    \item the storage of printed templates of all original printed codes requires expensive IT infrastructures;
    \item the online authentication is prevented, as a central defender system that uses original printed templates would be needed to avoid leakage of sensitive information.
\end{itemize}

Furthermore, it might be interesting to know the regions in CDP contributing the most to the authentication. In general, the anomaly localization in CDP represents an interesting tool for fake analysis.

In this respect, in this work, we propose a ML-based anomaly localization method that is based on digital templates only. The idea is to use the digital templates to estimate, through ML, the printed template that the defender would have obtained after printing. This allows us to base our authentication on digital templates only while retaining better performances with respect to a direct comparison. The system only requires digital templates and original printed templates at training stage. At the same time, no knowledge of fakes is required. That is why 
 we refer to it as unsupervised.

Our contributions are the following:
\begin{itemize}
    \item we propose an unsupervised anomaly localization system that performs authentication based on digital templates only and with no knowledge about fakes at the training stage;
    \item we perform the empirical evaluation of the proposed approach on the dataset of CDP designed to mimic real-life circumstances.
\end{itemize}

\noindent \textbf{Notations}  We use the following notations: $\mathcal{D}$ and $\mathcal{A}$ are the sets of printing processes available to the defender and attackers respectively; $\vect{t}_i \in \{0, 1\}^{H \times W}$ denotes the $i$-th original digital template; $\vect{x}_i^d \in [0, 1]^{H \times W}$ corresponds to the $i$-th original printed template printed using $d \in \mathcal{D}$, while $\vect{f}_i^{a/d} \in [0, 1]^{H \times W}$ is used to denote the respective printed fake code which template was estimated based on $\vect{x}_i^d$ and printed using process $a \in \mathcal{A}$; $\vect{y}_i \in [0, 1]^{H \times W}$ stands for a probe which might be either original or fake.
% ============================================================
\section{Problem formulation}
A defender, to protect its products from counterfeiting, generates a set of digital templates $\vect{t}_i \in \{0, 1\}^{H \times W}$ and prints them obtaining the respective original printed codes $\vect{x}_i^d \in [0, 1]^{H \times W}$, where $d$ identifies the used defender's printer and $d \in \mathcal{D}$, where $\mathcal{D}$ is the set of all printing processes available to the defender and $i$ denotes the identifier of the object. The printed original CDP are distributed in the public domain jointly with the objects being protected.

An attacker having an access to these publicly available codes scans them and estimates the digital templates $\vect{t}_i$ as $\tilde{\vect{t}}_i$. The obtained estimations are then printed obtaining thus fake printed copies $\vect{f}_i^{a/d}$, where $a \in \mathcal{A}$ indicates the printer used by the attacker and $d \in \mathcal{D}$ the printer used for printing the copied original CDP $\vect{x}_i^d$ by the defender. The fake CDP fabricated by the attacker are also put into the public domain. 

At inference time, which represents the authentication stage, given a probe $\vect{y}_i$ which could either be an original $\vect{x}_i^d$ or fake $\vect{f}_i^{a/d}$, the authentication system has to determine whether $\vect{y}_i$ is an original, i.e., $\vect{y}_i \in \{\vect{x}_i^{d} | d \in \mathcal{D}\}$, or fake, i.e., $\vect{y}_i \in \{\vect{f}_i^{a/d} | d \in \mathcal{D} \land a \in \mathcal{A}\}$. Notice that while referring to $\vect{x}_i^d$, $\vect{y}_i$, $\vect{f}_i^{a/d}$ we assume images acquired from the physical objects based on specified imaging devices.

In our work, the defender determines the nature of $\vect{y}_i$ through an anomaly map $a_{map}(\vect{t}_i, \vect{y}_i) \in [0, 1]^{H \times W}$ which highlights anomalous locations on $\vect{y}_i$.

%The authentication might be performed in different ways, in supervised, semi-supervised or unsupervised manner. In our work we propose and investigate the unsupervised approach based on the estimation of local anomalies in the fakes codes.
% ============================================================
\section{Authentication based on digital templates and printed templates}\label{sec:t_vs_x}
% --------------------------------
    \begin{figure}
        \centering
        \begin{subfigure}[b]{0.22\textwidth}
            \centering
            \includegraphics[width=\textwidth]{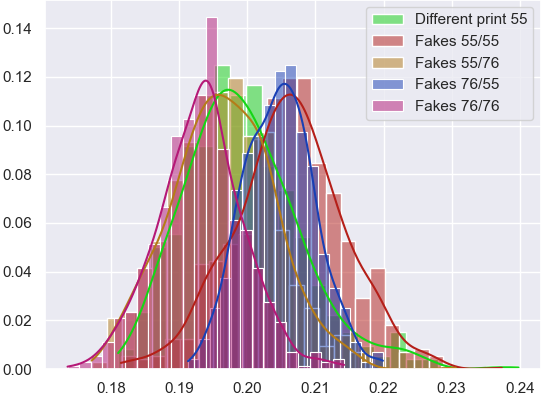}
            \caption{MSE w.r.t. digital templates highlighting different print of originals $\vect{x}^{55}$. Total AUC = $0.56$.} 
            \label{fig:sub_a}
        \end{subfigure}
        \hfill
        \begin{subfigure}[b]{0.22\textwidth}  
            \centering 
            \includegraphics[width=\textwidth]{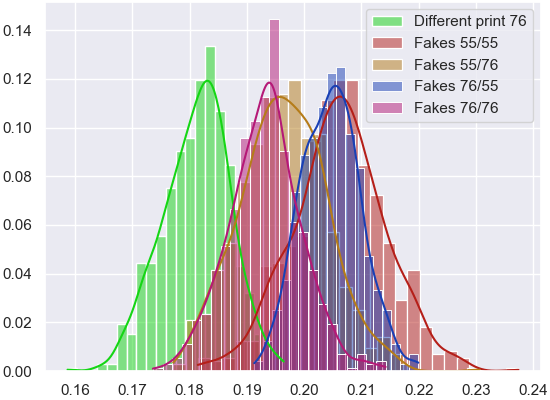}
            \caption{MSE w.r.t. digital templates highlighting different print of originals $\vect{x}^{76}$. Total AUC = $0.95$.} 
            \label{fig:sub_b}
        \end{subfigure}
        \vskip\baselineskip
        \begin{subfigure}[b]{0.22\textwidth}   
            \centering 
            \includegraphics[width=\textwidth]{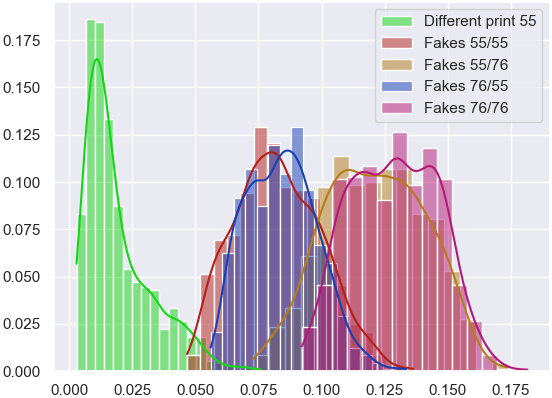}
            \caption{MSE w.r.t. printed templates $\vect{x}^{55}$ highlighting the different print of originals. Total AUC = $0.99$.}
            \label{fig:sub_c}
        \end{subfigure}
        \hfill
        \begin{subfigure}[b]{0.22\textwidth}   
            \centering 
            \includegraphics[width=\textwidth]{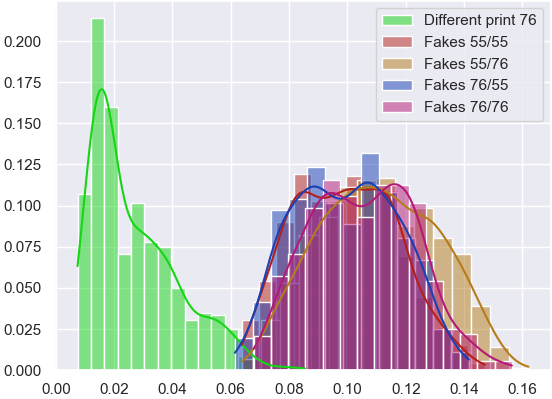}
            \caption{MSE w.r.t. printed templates $\vect{x}^{76}$ highlighting the different print of originals. Total AUC = $0.99$.}
            \label{fig:sub_d}
        \end{subfigure}
        \caption{Histograms of MSE for original and fake codes with respect to digital templates (\autoref{fig:sub_a} and \autoref{fig:sub_b}) and printed templates (\autoref{fig:sub_c} and \autoref{fig:sub_d}).} 
        \label{fig:comparison}
    \end{figure}
% --------------------------------

In our study, we focus on the CDP authentication facing the ML attacks \cite{romanwifs21} that are shown to be a real challenge for the authentication system based on the digital templates. Simple similarity metrics such as Pearson correlation coefficient \cite{picard} or mean squared error (MSE) can not reliably differentiate the originals from the ML fakes based on a test $d_{sim}(\vect{y}_i, \vect{t}_i) \le \gamma$, where $d_{sim}(\cdot)$ denotes a similarity metric and $\gamma$ stands for the threshold. At the same time, as mentioned in \autoref{sec:intro}, the supervised deep classifiers are subject to distribution shift when facing unseen fakes \cite{icip22}. To demonstrate the inability of the above test to deal with the ML fakes, we have used the Indigo 1x1 base dataset of originals and fakes printed on two industrial printers HP Indigo 5500 DS (denoted as 55) and HP Indigo 7600 DS (denoted as 76) from \cite{romanwifs21}. In \autoref{fig:sub_a} and \autoref{fig:sub_b} we plot the MSE between the digital templates and fakes and highlighting (in green) the MSE between digital templates and originals 55 and 76, respectively. 

We confirm that the authentication based on digital templates is less accurate than authentication based on printed templates when considering the simple MSE metric. Such printed templates are images acquired at the enrollment stage from the physical objects. \autoref{fig:sub_c} and \autoref{fig:sub_d} show the corresponding statistics. The histograms of scores for original CDP and fakes are much more distinctive. The area under the curve (AUC) score also confirms the superior performance of printed template-based authentication for both printers.

% -------------------------
\begin{figure*}[t]
    \centering
    \includegraphics[width=0.95\linewidth]{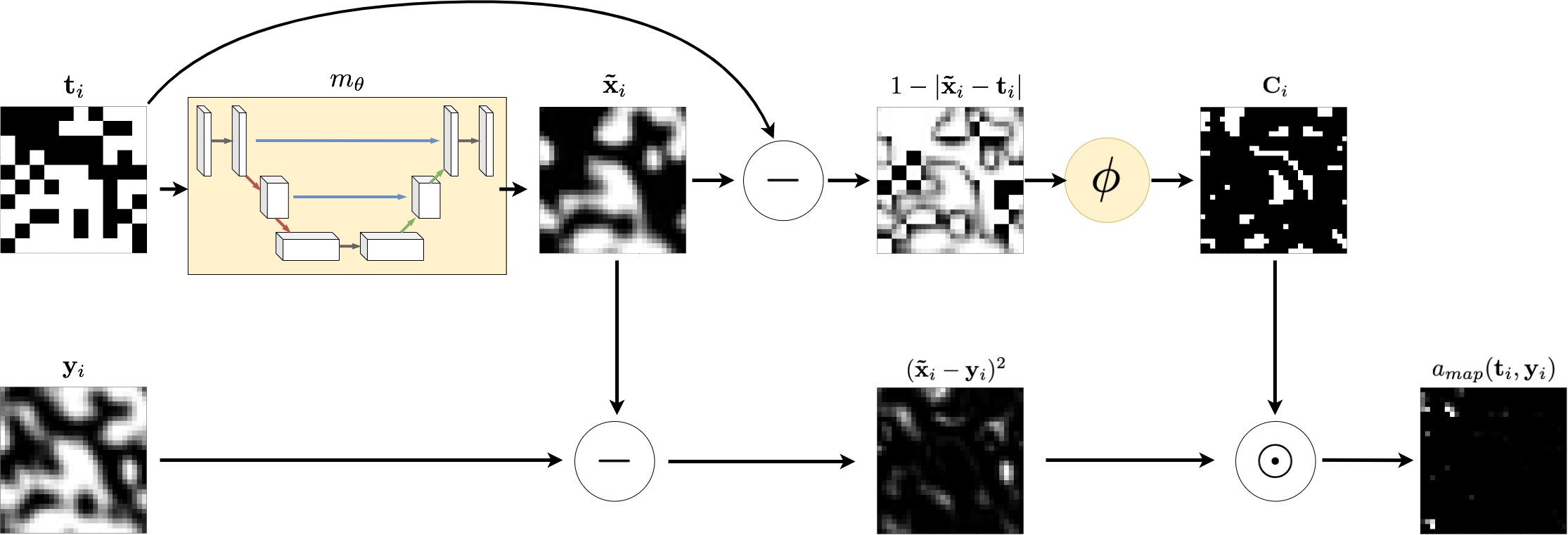}
    \caption{Proposed anomaly localization method: probe $\vect{y}_i$, which could be an original or fake code, is compared with an estimate $\vect{\tilde{x}}_i$ of original code $\vect{x}_i$ based on the digital template $\vect{t}_i$. This comparison is weighted by a confidence map $\vect{C}_i$, which captures the uncertainty on the outcome of the  defender's print process. The result is an anomaly map where brightest pixels highlight differences with respect to the estimate $\tilde{\vect{x}}_i$ that are anomalous.}
    \label{fig:method}
\end{figure*}
% ----------------------------

This result is due to the fact that authentication based only on digital templates heavily relies on the ability of the printer to accurately reproduce the digital template structure of the CDP. We are aware that printer 55, which is an industrial printer just like 76, produces more distortions and this is reflected in the fact that authentication based on digital templates performs much worse for such printed originals (\autoref{fig:sub_a}) than those printed with printer 76 (\autoref{fig:sub_b}). 

However, the authentication based on the printed templates is not influenced by the gap between the digital template and the printed codes. Therefore, the distance between the probe represented by the authentic CDP and the printed template is minimized to the acquisition distortion and the impact of the printing distortions is not so relevant for the defender. In contrast, the printing distortions play an important role for the attacker on the way toward an accurate estimation of the digital template from the acquired CDP.

In our work, we propose a system that performs the authentication similarly to how it is done for physical templates while only requiring digital templates, thus solving the impractical difficulties introduced in \autoref{sec:intro}.

\section{Proposed framework}\label{sec:method}
Motivated by our findings in \autoref{sec:t_vs_x}, we create a ML model which is capable, given a digital template, to estimate the respective printed template and use this estimation to localize anomalies. This approach makes sure that authentication is based on digital templates only, but tries to achieve better performances similar to those obtained when authenticating based on printed templates. We present our method schematically in \autoref{fig:method}.

% --------------------------------
\begin{figure*}[t]
    \centering
    \begin{subfigure}[b]{0.42\textwidth}
        \centering
        \includegraphics[width=\textwidth]{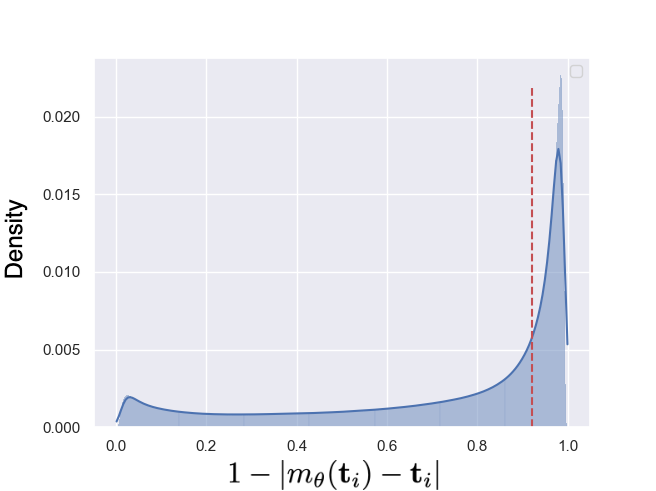}
        \caption{Agreement between template and print estimation for system trained on originals $\vect{x}^{55}$.}
        \label{fig:hist_err_55}
    \end{subfigure}
    \hfill
    \begin{subfigure}[b]{0.42\textwidth}
        \centering
        \includegraphics[width=\textwidth]{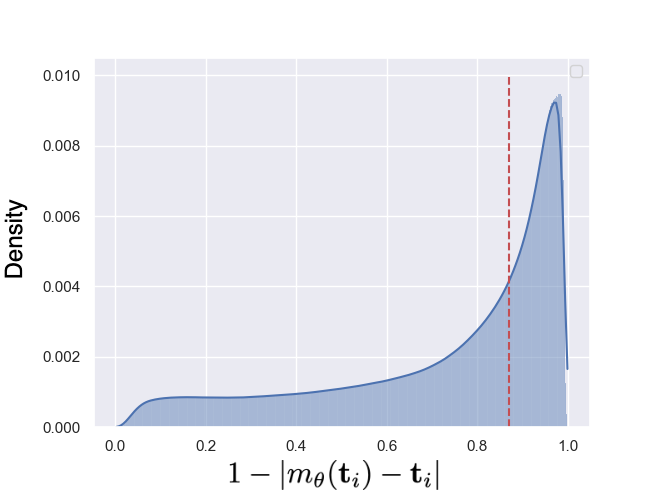}
        \caption{Agreement between template and print estimation for system trained on originals $\vect{x}^{76}$.}
        \label{fig:hist_err_76}
    \end{subfigure}
    \caption{Histograms showing weight $1 - |m_\theta(\vect{t}_i) - \vect{t}_i|$ at each pixel location for models trained for both printers using a subset of 100 codes. Red dashed lines denote the imposed threshold.}
    \label{fig:hists_err}
\end{figure*}
% ---------------------------------

Given a paired dataset of digital templates $\vect{t}_i$ and respective printed original CDP $\vect{x}_i^d$, we learn a model $m_{\theta}$, parametrized by learnable parameters $\theta$, which imitates the specific printing process $d$ such that $m_{\theta}(\vect{t}_i) = \tilde{\vect{x}}_i \approx \vect{x}_i^d \ \forall i$. This model can, in principle, be any image-to-image model.

% At the authentication stage, given a probe $\vect{y}_i$, we obtain its anomaly map as follows:

The authentication of a probe $\vect{y}_i$ is performed based on its anomaly map defined as:

\begin{equation}
a_{map}(\vect{t}_i, \vect{y}_i) = \vect{C}_i \odot (m_\theta(\vect{t}_i) - \vect{y}_i)^2,
\end{equation}

where $\odot$ represents the element-wise product and $\vect{C}_i$ represents a measure of the confidence held by the defender on the value of each pixel in $\vect{x}_i^d$ with respect to its digital template:

\begin{equation}
\vect{C}_i = \phi(\mathbf{1} - | \vect{t}_i - m_\theta(\vect{t}_i) |),
\end{equation}

where $\mathbf{1}$ is a matrix of ones and $\phi(\cdot)$ is any element-wise increasing function s.t. $\phi(0) = 0$ and $\phi(1) = 1$ (e.g, exponential, masking by a threshold, etc.). Intuitively, function $\phi(\cdot)$ serves as a weighting function and ensures that when the difference $|\vect{t}_i - m_\theta(\vect{t}_i)|$ is relatively high with respect to all pixels for all codes in the training set, the confidence is diminished accordingly. Note that $a_{map}(\vect{t}, \vect{y}) \in [0, 1]^{H \times W}$.

Also, note that we assume the knowledge of the associated digital template $\vect{t}_i$ given a probe $\vect{y}_i$ since it is straightforward to recover the most similar template given a printed code.

The anomaly map $a_{map}(\vect{t}_i, \vect{y}_i)$ can then be used to exactly locate anomalies as well as to assign an anomaly score $a_{score}(\vect{t}_i, \vect{y}_i)$ to the printed code through some aggregation function $s(\cdot)$:

\begin{equation}
    a_{score}(\vect{t}_i, \vect{y}_i) = s(a_{map}(\vect{t}_i, \vect{y}_i)),
\end{equation}

which could be for example $\ell_1$-norm, $\ell_2$-norm, etc. The anomaly score is used for anomaly detection: the defender can set a threshold $\gamma$ such that the probe $\vect{y}_i$ is labeled as anomalous if $a_{score}(\vect{y}_i) > \gamma$ and as non-anomalous otherwise.

We define the proposed method as unsupervised as it does not rely on fake codes, but can be trained from a paired set of digital templates and original printed CDP. Furthermore, the proposed system does not require printed templates for the authentication of $\vect{y}_i$. Finally, such a system can also be used in the case where the defender disposes of multiple printing processes and a model $m_\theta$ is learned for each of such printing processes.

% --------------------------------

\begin{table*}[t!]\centering
\ra{1.2}
\begin{tabular}{@{}rrrrcrrrcrrr@{}}\toprule
 & \multicolumn{4}{c}{$\vect{x}^{55}$} & \phantom{abc} & \multicolumn{4}{c}{$\vect{x}^{76}$} \\
\cmidrule{2-5}  \cmidrule{7-10}
& $\vect{f}^{55/55}$ & $\vect{f}^{55/76}$ & $\vect{f}^{76/55}$ & $\vect{f}^{76/76}$ && $\vect{f}^{55/55}$ & $\vect{f}^{55/76}$ & $\vect{f}^{76/55}$ & $\vect{f}^{76/76}$\\
Supervised trained on $\vect{x}^{55}, \vect{f}^{55/55}$ \cite{icip22} & 1.00 & - & 1.00 & - && - & - & - & - \\
Supervised trained on $\vect{x}^{55}, \vect{f}^{76/55}$ \cite{icip22} & 0.70 & - & 1.00 & - && - & - & - & - \\
Supervised trained on $\vect{x}^{76}, \vect{f}^{55/76}$ \cite{icip22} & - & - & - & - && - & 1.00 & - & 0.26 \\
Supervised trained on $\vect{x}^{76}, \vect{f}^{76/76}$ \cite{icip22} & - & - & - & - && - & 0.25 & - & 1.00 \\
$MSE(\vect{t}, \vect{y})$ & 0.73 & 0.41 & 0.72 & 0.28 && \textbf{0.99} & 0.94 & \textbf{0.99} & 0.92 \\
$a_{score}(\vect{t}, \vect{y})$ without $\vect{C}_i$ & 0.98 & 0.98 & \textbf{1.00} & \textbf{1.00} && 0.98 & 0.98 & \textbf{0.99} & \textbf{0.99} \\
$a_{score}(\vect{t}, \vect{y})$ with $\vect{C}_i$ & \textbf{0.99} & \textbf{0.99} & \textbf{1.00} & \textbf{1.00} && \textbf{0.99} & \textbf{0.99} & \textbf{0.99} & \textbf{0.99} \\
$MSE(\vect{x}, \vect{y})$ & \textbf{0.99} & \textbf{0.99} & 0.99 & \textbf{1.00}  && \textbf{0.99} & \textbf{0.99} & \textbf{0.99} & \textbf{0.99} \\
\bottomrule
\end{tabular}
\caption{AUC scores using the supervised systems \cite{icip22}, the MSE metrics and our proposed system. The mean AUC score over 10 runs with different randomizing seeds are shown.}
\label{tab:aucs}
\end{table*}

% -----------------------------------
\begin{figure*}[!t]
    \centering
    \begin{subfigure}[b]{0.42\textwidth}
        \centering
        \includegraphics[width=\textwidth]{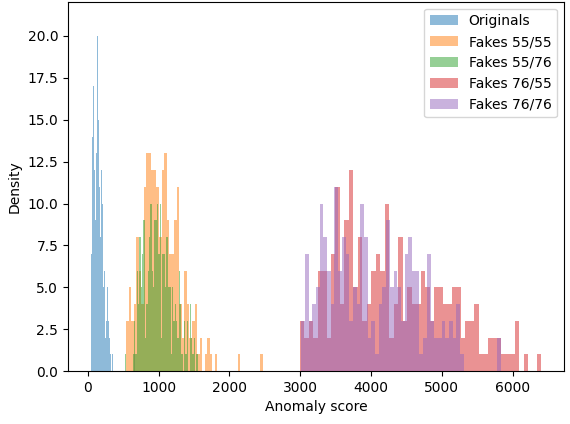}
        \caption{Histogram of anomaly scores for system trained on $\vect{x}^{55}$}
        \label{fig:hist55}
    \end{subfigure}
    \hfill
    \begin{subfigure}[b]{0.42\textwidth}
        \centering
        \includegraphics[width=\textwidth]{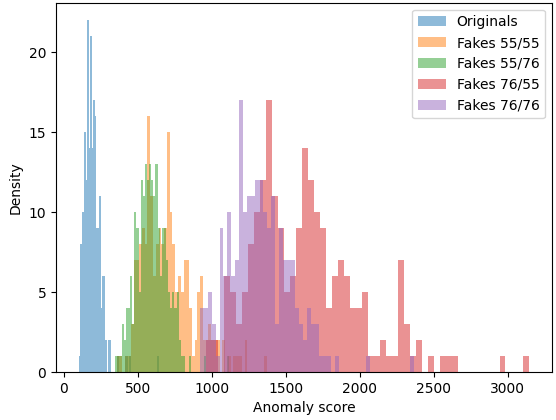}
        \caption{Histogram of anomaly scores for system trained on $\vect{x}^{76}$}
        \label{fig:hist76}
    \end{subfigure}
    \caption{Histograms of anomaly scores with original and fake CDP for one run.}
    \label{fig:hists}
\end{figure*}
% ---------------------------------------

\section{Experiments}
\subsection{Experimental setup}
To evaluate the proposed system, we study its anomaly detection capabilities based on the extracted anomaly maps. We run our experiments on the Indigo 1x1 base dataset presented in \cite{romanwifs21} and publicly available at \cite{dataset}, which is composed of a set of digital templates, two sets of printed originals, and four sets of printed fake CDP. Each set contains 720 CDP of size $684 \times 684$ pixels. The two sets of original codes were obtained by printing the set of digital templates onto two distinct industrial printers, namely HP Indigo 5500 DS (55) and HP Indigo 7600 DS (76). Each of the four sets of fake CDP was obtained in the following manner: given a set of printed original CDP, a ML model is used to predict the corresponding digital templates and the obtained estimations are then printed on either the same printer or a different one.

The four sets of fake CDP vary on the original printed CDP used to estimate the digital templates and on the used printer. We denote the printers used for original and fake CDP with the superscripts $55$ and $76$, respectively.

We train a print-imaging model $m_\theta$ for each set of original CDP at our disposal, thus resulting in two separate authentication systems.

For our experiments, we set the aggregation function $s(\cdot)$ to be the summation of all values in $a_{map}(\vect{t}_i, \vect{y}_i)$, whereas for $\phi(\cdot)$ we use a function that simply sets to zero all values below a given threshold. Threshold values for originals 55 and 76 differ. Finally, we adopt a shallow U-Net-like architecture \cite{unet} for $m_\theta$, which we train using a 60/10/30\% training-validation-test split, the MSE loss function, and the Adam optimizer with a learning rate of $10^{-2}$.

\subsection{Stochasticity of printing}
By training models $m_\theta$ for printers 55 and 76 and varying the architecture, we found that a ML model is incapable of perfectly imitating the printing process. In \autoref{fig:hists_err}, we display the distribution of weight $1 - |m_\theta(\vect{t}_i) - \vect{t}_i|$ for both printers using a corresponding subset of 100 digital templates. The plots highlight how despite most of the pixels being correctly predicted by the trained model (values close to $1$), a good part of them remains poorly approximated due to the stochastic nature of the printing processes. While the performances obtained with stochastic models would need investigation, we only train deterministic models $m_\theta$ because of the lack of a multitude of printed originals given the same digital template in our dataset.

\subsection{Role of the confidence map}\label{subsec:confidence}
Given that printing is a stochastic process, our intuition is that we should not look for anomalies at pixel locations where the pre-trained model $m_\theta$ is incapable of predicting the outcome of the print process. Instead, we should only consider pixel locations where the measure $1 - |m_\theta(\vect{t}_i) - \vect{t}_i|$ is above a certain threshold while using as many reliable pixels as possible for the authentication.

By studying \autoref{fig:hists_err}, we empirically find that threshold values of 0.87 for $m_\theta$ trained on originals $55$ and of 0.92 for $m_\theta$ trained on originals $76$ seem good trade-offs between the ratio of the number of pixels used versus confidence in their values. Setting $\phi(\cdot)$ to be functions which threshold on the aforementioned values allow us to use $41\%$ and $43\%$ of the pixels for originals $55$ and $76$ in the sample, respectively. This empirical choice has a small impact on the overall performance of the system, but we find that weighting the MSE between $\tilde{\vect{x}}_i$ and $\vect{y}_i$ by the obtained confidence $\vect{C}_i$ enhances the overall performances slightly in the considered setup and available dataset.

\subsection{Results}
% ---------------------------------------
\begin{figure}[t!]
    \centering
    \begin{subfigure}[b]{0.45\textwidth}
        \centering
        \includegraphics[width=\textwidth]{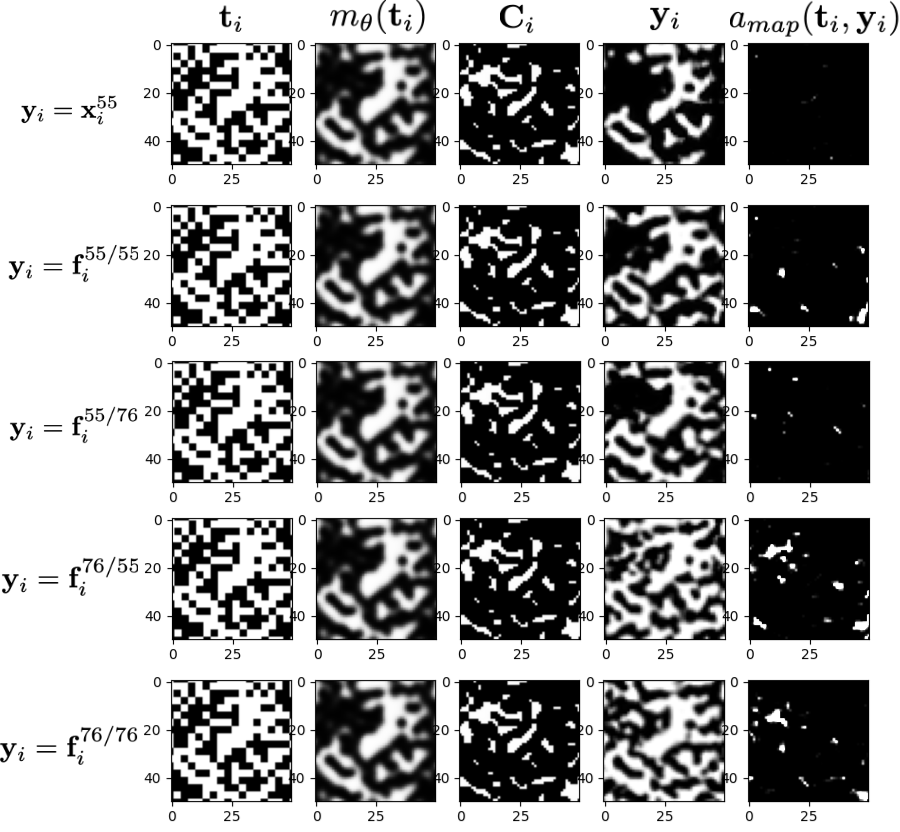}
        \caption{System trained on $\vect{x}^{55}$}
        \label{fig:codes55}
    \end{subfigure}
    
    \vspace{0.1\textwidth}
    
    \begin{subfigure}[b]{0.45\textwidth}
        \centering
        \includegraphics[width=\textwidth]{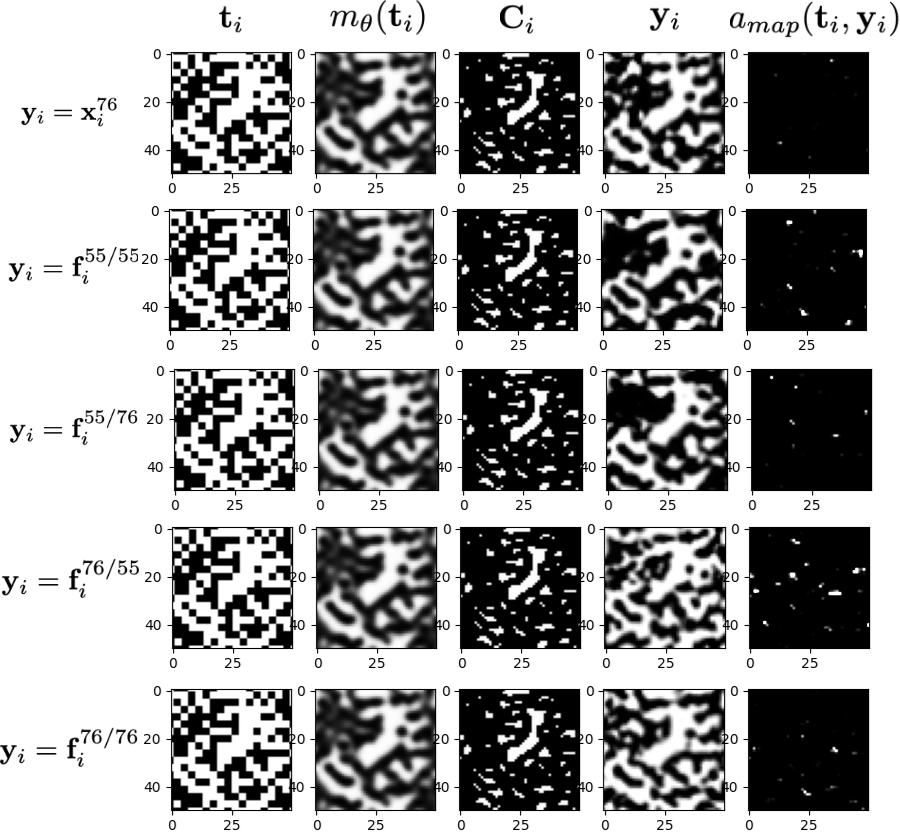}
        \caption{System trained on $\vect{x}^{76}$}
        \label{fig:codes76}
    \end{subfigure}
    \caption{30x30 crop of templates $\vect{t}_i$, synthetic print estimations $m_\theta(\vect{t}_i) = \tilde{\vect{x}}_i$, confidence map $\vect{C}_i$, test probes $\vect{y}_i$ and anomaly maps $a_{map}(\vect{t}_i, \vect{y}_i)$ for system trained on originals 55 (a) and 76 (b). Brighter spots on confidence and anomaly maps represent higher confidence and anomaly, respectively.}
    \label{fig:codes}
\end{figure}
% ----------------------------------------

We validate the performances of the proposed systems against all types of fake CDP. In \autoref{tab:aucs} we report the mean of the AUC-score that the systems achieved over 10 distinct runs with different randomizing seeds. We compare our methodology against authentications presented in \autoref{sec:t_vs_x}, namely the simple MSE metric using digital templates and different printed templates. We also include the results obtained in \cite{icip22} for supervised systems trained on one type of original and fake CDP.

We clearly see in \autoref{tab:aucs} that a supervised system, generally, performs very poorly on fakes that have not been used for training and is not robust to a distribution shift. As anticipated in \autoref{sec:t_vs_x}, MSE between printed templates and test probes allows better separability of originals and fakes than using the same measure with digital templates. Due to lower printing precision, results in authentication based on digital templates for printed 55 are worse than for printer 76. Our method, which benefits from the advantages of both digital and printed template based authentications, achieves results that are very close to the authentication based on printed templates.

In \autoref{fig:hists} we show the histogram of anomaly scores for original and fake CDP for the trained systems for both printers.

We find our method to be capable, similarly to authentication based on printed templates, to distinguish originals from fakes with high accuracy for all possible combinations of original and fake CDP over different runs. Furthermore, since nearly perfect separability is achieved, the defender can set the threshold $\gamma$ as:
\begin{equation}
    \gamma = \max_{i} \ a_{score}(\vect{t}_i, \vect{x}_i)
\end{equation}
and be certain to not miss any original while rejecting almost entirely fake CDP.

Finally, we show examples of the obtained results in \autoref{fig:codes}. For both sub-figures, the first three columns are static and show the digital template $\vect{t}_i$, the expected printed code $m_\theta(\vect{t}_i)=\tilde{\vect{x}}_i$ and the confidence $\vect{C}_i$. The fourth and fifth column of both sub-figures show the test probe $\vect{y}_i$ and the obtained anomaly map $a_{map}(\vect{t}_i, \vect{y}_i)$, respectively. The first row shows the case when the test probe is the original CDP, whereas the remaining four rows show the cases when the test probe is one of the aforementioned types of fake CDP. The figure clearly shows that the anomaly maps for original codes are much less active (lack white spots) than those obtained with fake CDP, where anomalies are detected. Moreover, the anomaly map clearly show the regions of largest anomalies, thus demonstrating the anomaly localization capacity of the proposed system.

\newpage
\section{Conclusion}
In this work, we proposed an authentication system for CDP which can localize anomalies based on digital templates only, while only requiring original digital and printed CDP for training. We trained a deterministic ML model to imitate the print-imaging process of the defender and used it for comparison against test probes. This comparison is weighted by a measure of confidence, which reduces the importance of detected differences based on the stochasticity of the print process at such locations.

We evaluated our system on the task of anomaly detection, where the anomaly maps were reduced to anomaly scores used to compute the AUC score. This resulted in nearly perfect separability between original and fake CDPs.

For future work, we aim at investigating the performance of the proposed system with respect to CDP acquired with mobile phones under enrollment settings close to real-life conditions. In addition, we aim at investigating the performance of a model that mimics the stochasticity of the printing process and thus produces small print deviations for the same digital template.

\bibliographystyle{./bibstyle}
\bibliography{biblio}

\begin{thebibliography}{1}
\providecommand{\url}[1]{#1}
\csname url@samestyle\endcsname
\providecommand{\newblock}{\relax}
\providecommand{\bibinfo}[2]{#2}
\providecommand{\BIBentrySTDinterwordspacing}{\spaceskip=0pt\relax}
\providecommand{\BIBentryALTinterwordstretchfactor}{4}
\providecommand{\BIBentryALTinterwordspacing}{\spaceskip=\fontdimen2\font plus
\BIBentryALTinterwordstretchfactor\fontdimen3\font minus
  \fontdimen4\font\relax}
\providecommand{\BIBforeignlanguage}[2]{{%
\expandafter\ifx\csname l@#1\endcsname\relax
\typeout{** WARNING: IEEEtran.bst: No hyphenation pattern has been}%
\typeout{** loaded for the language `#1'. Using the pattern for}%
\typeout{** the default language instead.}%
\else
\language=\csname l@#1\endcsname
\fi
#2}}
\providecommand{\BIBdecl}{\relax}
\BIBdecl

\bibitem{picard}
J.~Picard, ``Digital authentication with copy-detection patterns,''
  \emph{Electron. Imaging}, vol. 5310, 06 2004.

\bibitem{voloshynovskiy2016physical}
S.~Voloshynovskiy, T.~Holotyak, and P.~Bas, ``Physical object authentication:
  Detection-theoretic comparison of natural and artificial randomness,'' in
  \emph{2016 IEEE International Conference on Acoustics, Speech and Signal
  Processing (ICASSP)}.\hskip 1em plus 0.5em minus 0.4em\relax IEEE, 2016, pp.
  2029--2033.

\bibitem{Taran2020icassp}
O.~Taran, S.~Bonev, T.~Holotyak, and S.~Voloshynovskiy, ``Adversarial detection
  of counterfeited printable graphical codes: towards "adversarial games" in
  physical world,'' in \emph{IEEE International Conference on Acoustics, Speech
  and Signal Processing (ICASSP)}, 2020.

\bibitem{romanwifs21}
R.~Chaban, O.~Taran, J.~Tutt, T.~Holotyak, S.~Bonev, and S.~Voloshynovskiy,
  ``Machine learning attack on copy detection patterns: are 1x1 patterns
  cloneable?'' in \emph{IEEE International Workshop on Information Forensics
  and Security (WIFS)}, Montpellier, France, December 2021.

\bibitem{khermaza2021can}
E.~Khermaza, I.~Tkachenko, and J.~Picard, ``Can copy detection patterns be
  copied? evaluating the performance of attacks and highlighting the role of
  the detector,'' in \emph{2021 IEEE International Workshop on Information
  Forensics and Security (WIFS)}.\hskip 1em plus 0.5em minus 0.4em\relax IEEE,
  2021, pp. 1--6.

\bibitem{icip22}
B.~Pulfer, R.~Chaban, Y.~Belousov, J.~Tutt, O.~Taran, T.~Holotyak, and
  S.~Voloshynovskiy, ``Authentication of copy detection patterns under machine
  learning attacks: A supervised approach,'' in \emph{IEEE International
  Conference on Image Processing (ICIP)}, Bordeaux, France, October 2022.

\bibitem{dataset}
R.~Chaban and O.~Taran, ``{Indigo 1x1 base},''
  \url{https://github.com/sip-group/snf-it-dis/tree/master/datasets/indigo1x1base},
  2021.

\bibitem{unet}
\BIBentryALTinterwordspacing
O.~Ronneberger, P.~Fischer, and T.~Brox, ``U-net: Convolutional networks for
  biomedical image segmentation,'' 2015. [Online]. Available:
  \url{https://arxiv.org/abs/1505.04597}
\BIBentrySTDinterwordspacing

\end{thebibliography}

\end{document}